\documentclass[letterpaper, 10 pt, conference]{ieeeconf}      
\usepackage{amssymb,amsmath,graphicx,color,algorithm,algorithmic,url,array}
\usepackage[english]{babel}
\usepackage[noadjust]{cite} 
\usepackage{hyperref}

\IEEEoverridecommandlockouts                              

\title{Towards integrated tactile sensorimotor control \\ in anthropomorphic soft robotic hands
\vspace{0em}
}

\author{Nathan F. Lepora, Andrew Stinchcombe, Chris Ford, Alfred Brown, John Lloyd, \\ Manuel G. Catalano, Matteo Bianchi, Benjamin Ward-Cherrier%
\thanks{$^{1}$NL, AS, CF, AB, JL and BWC are with the Department of Engineering Mathematics and Bristol Robotics Laboratory, University of Bristol, U.K.}%
\thanks{$^{2}$MBC is with the Centro di Ricerca E. Piaggio e Dipartimento di Ingegneria
	dell'Informazione, Universit di Pisa, Pisa, Italia}%
\thanks{$^{3}$MGB is with Istituto Italiano di Tecnologia, Genova, Italia}%
\thanks{Corresponding author: n.lepora@bristol.ac.uk}%
}

\begin{document}

\maketitle

\begin{abstract}
In this work, we report on the integrated sensorimotor control of the Pisa/IIT SoftHand, an anthropomorphic soft robot hand designed around the principle of adaptive synergies, with the BRL tactile fingertip (TacTip), a soft biomimetic optical tactile sensor based on the human sense of touch. Our focus is how a sense of touch can be used to control an anthropomorphic hand with one degree of actuation, based on an integration that respects the hand's mechanical functionality. We consider: (i)~closed-loop tactile control to establish a light contact on an unknown held object, based on the structural similarity with an undeformed tactile image; and (ii) controlling the estimated pose of an edge feature of a held object, using a convolutional neural network approach developed for controlling other sensors in the TacTip family. Overall, this gives a foundation to endow soft robotic hands with human-like touch, with implications for autonomous grasping, manipulation, human-robot interaction and prosthetics. 
\end{abstract}


\section{INTRODUCTION}

Replication of the human hand and its functionality is one of the major goals of robotics, with expected widespread industrial and societal ramifications. There are emerging trends in the design of robot hands, such as soft actuation and the use of underactuation~\cite{piazza_century_2019,birglen_underactuated_2007,odhner_compliant_2014,deimel_novel_2016,pozzi_hand_2020}, which have a close link with biological principles of motor control. The versatility and execution of the human hand depends on a coupling between its mechanical structure and its sensory capabilities in the form of proprioceptive and tactile feedback. How these sensorimotor capabilities interact to give humans their unique dexterity is not understood yet. However, it seems reasonable to consider them as core capabilities whose embodiment in robot hands could bridge the gap between human and robot performance in autonomous grasping and manipulation.


As emphasised recently in a review of the trends and challenges in robot manipulation~\cite{billard_trends_2019}, it remains an open problem to construct robotic hands that contain all of the mechanical and haptic sensing components necessary to control the hand to grasp, manipulate and explore objects with the ease of the human hand. The present work presents an integrated sensorimotor control of an anthropomorphic hand based on two complementary soft robotic technologies: the Pisa/IIT SoftHand~\cite{catalano_adaptive_2014} and BRL tactile fingertip (TacTip)~\cite{ward-cherrier_tactip_2018,chorley_development_2009}. An initial integration was proposed in~\cite{ward-cherrier_miniaturised_2020}, which is progressed here to implement the first sensorimotor control loop of a SoftHand using this unique combined technology.


The SoftHand is an anthropomorphic soft articulated robot hand whose design and control are based on postural synergies that represent a reduced set of principal directions in hand configuration space describing the most frequent postures in human hand movements~\cite{santello_hand_2016}. Endowing soft hands with advanced sensing capabilities comes with challenges (discussed in the next section), with several solutions proposed~\cite{santaera_low-cost_2015,bianchi_touch-based_2018,casini_design_2015,zoller_acoustic_2018}, but none has captured the rich geometric information from optical tactile sensors such as the MIT GelSight~\cite{yuan_gelsight_2017,johnson_retrographic_2009} or BRL TacTip~\cite{ward-cherrier_tactip_2018,chorley_development_2009}. Of these, the TacTip is based on the dermal papillae structure in human tactile skin where low-threshold mechanoreceptors are localized~\cite[Fig.1]{abraira_sensory_2013}, which are fabricated by 3D-printing pin-like structures in a compliant skin imaged with a camera~\cite{ward-cherrier_tactip_2018}. Given the soft biomimetic nature of the SoftHand, the TacTip offers a complementary soft biomimetic sense of touch.

\begin{figure}[t]
	\centering
	\begin{tabular}{@{}c@{}}
		\includegraphics[width=1\columnwidth,trim={50 20 135 20},clip]{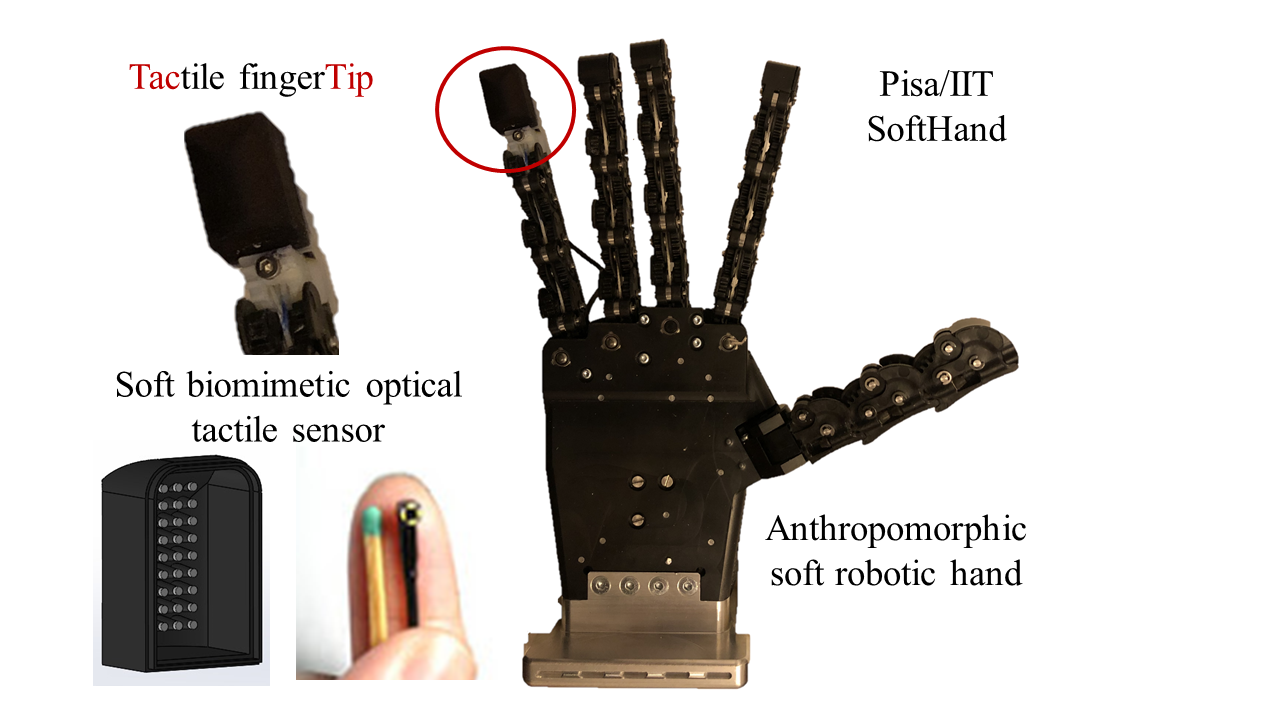}
	\end{tabular}
	\caption{The BRL biomimetic optical tactile fingertip (TacTip) integrated~as the distal phalanx of the fifth digit of the anthropomorphic Pisa/IIT SoftHand} 
	\vspace{-1.25em}
	\label{fig1}	
\end{figure}

In this paper, an integration of the Pisa/IIT SoftHand and BRL TacTip is used to investigate sensorimotor control based on the tactile feedback. First, we demonstrate closed-loop tactile control to maintain delicate hand closure on an unknown object, using a basic measure of contact deformation with the Structural Similarity Index Measure (SSIM) against an undeformed tactile image. Second, we validate that the customized tactile sensor can accurately perceive the pose of edge features of held objects, aiming to emulate the recent performance of the BRL TacTip with deep learning~\cite{lepora_optimal_2020}. This pose-based control is then used to control the grasp closure and track the object pose. Success on these tests indicates the promise of this tactile hand for other sensorimotor tasks.


\section{BACKGROUND AND RELATED WORK}

Tactile sensing in dexterous robot hands has been a major focus of robotics research for decades, with a huge variety of different hand and sensor combinations~\cite{kappassov_tactile_2015}. In part, this proliferation has risen because of a lack of guiding principles to encourage roboticists along a fruitful path to tame the complexity of hands~\cite{bicchi_social_2015}. One approach, pursued here, is to exploit the principles underlying the function of the human hand to inform technological advances. Indeed, although the debate on the anthropomorphism in robotics is still open~\cite{bartneck_measurement_2009}, the human hand still represents the unmatched golden standard for dexterous manipulation and haptic exploration. 




The Pisa/IIT SoftHand combines two key principles that underlie human dexterity: postural synergies and soft actuation~\cite{santello_postural_1998}, which results in a five-fingered anthropomorphic robot hand that is simple to control yet moves with the primary motions of the human hand. The original design implemented one soft synergy~\cite{catalano_adaptive_2014}, resulting in a wide range of grasps controlled by just one degree of actuation. More recently, this has progressed to the Pisa/IIT SoftHand 2 by implementing a second postural synergy, which enables a broader range of grasps and some in-hand manipulation~\cite{santina_toward_2018}. 

Endowing soft robot hands with tactile sensing and proprioceptive capabilities is a challenging task~\cite{billard_trends_2019}. Classical solutions for rigid hands, {\em e.g.} joint encoders, cannot be applied in a straightforward manner to deformable structures. Solutions based on inertial measurements for hand posture reconstruction~\cite{santaera_low-cost_2015} or high frequency acceleration and contact detection~\cite{bianchi_touch-based_2018} have been proposed, alongside indirect estimation methods, such as measuring the motor current powering the artificial hand~\cite{casini_design_2015}. However, although promising, these techniques are still far from fully capturing the rich spectrum of tactile cues arising from the interaction with external objects in a manner resembling human touch.

The BRL TacTip is based on the principle that transduction in the upper layers of human skin takes place via the deformation of dermal papillae~\cite{ward-cherrier_tactip_2018,chorley_development_2009}. These structures enable sensing primarily via shear, providing an indirect measure of pressure~\cite{platkiewicz_haptic_2016}, unlike most artificial tactile sensors that use pressure-sensitive taxel arrays. Originally, the TacTip had a soft tactile dome of $40\,$mm diameter, which has since been customized into a family of biomimetic tactile sensors of various 3D-printed morphologies~\cite{ward-cherrier_tactip_2018} suitable both for stand-alone use and integration with robotic grippers large enough to accommodate the same USB camera~\cite{ward-cherrier_tactile_2016,ward-cherrier_model-free_2017,pestell_sense_2019}.

Until recently, an issue with optical tactile sensors is that they have needed to be large to accommodate a camera~\cite{ward-cherrier_tactip_2018,yuan_gelsight_2017}, which precluded sensors the size of a human fingertip. This problem is being overcome with the ongoing miniaturization of camera technology that has resulted in smaller versions of the GelSight integrated with robotic grippers~\cite{gomes_geltip_2020,romero_soft_2020,donlon_gelslim_2018,wilson_design_2020,lambeta_digit_2020} and a version of the TacTip integrated with the 3-fingered Model-O OpenHand~\cite{james_tactile_2020,james_slip_2020}. 

The integration of optical tactile sensing into 5-fingered anthropomorphic hands gives a new level of challenge to reach the size and shape of a human fingertip while respecting the hand's mechanical integrity. As far as we know, the only optical tactile sensor integrated with an anthropomorphic 5-fingered robot hand is a neuromorphic (event-based) version of the TacTip on the qb SoftHand~\cite{ward-cherrier_miniaturised_2020}. Here we introduce a smaller digital version of the TacTip on a Pisa/IIT SoftHand with a tendon layout that aids customization of the distal phalanx. By using a digital (rather than event-based) TacTip, we are able to utilize deep neural network methods that have been applied recently to controlling the TacTip~\cite{lepora_optimal_2020,lepora_pixels_2019}, which are here developed into an integrated approach for sensorimotor control of the SoftHand.




\section{METHODS}
\label{sec:3}

\begin{figure}[t]
	\centering
	\begin{tabular}{@{}c@{}}
		\includegraphics[width=1\columnwidth,trim={150 60 220 100},clip]{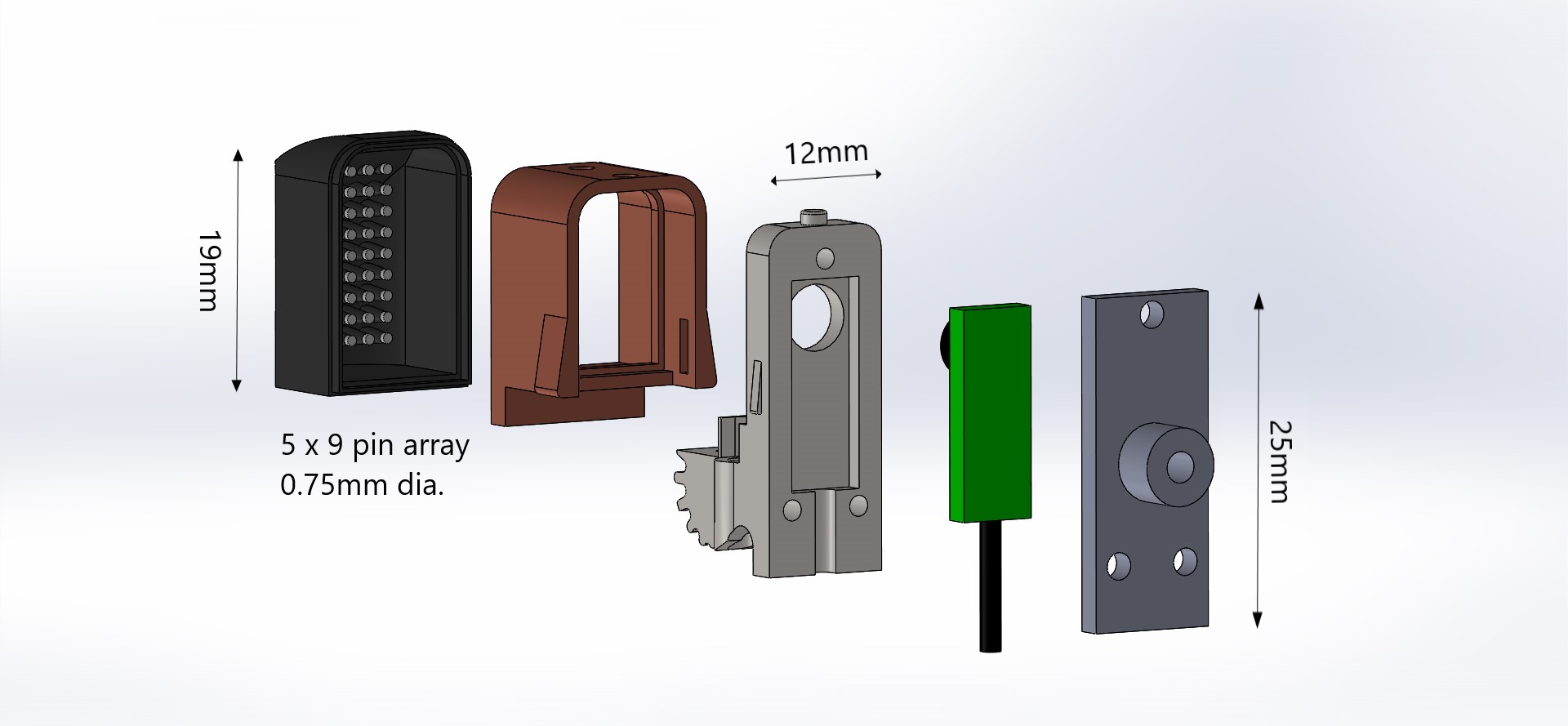}	
	\end{tabular}
	\caption{CAD of the tactile fingertip (approx 1:1 scale on page). The tip (to left; black and brown) is printed as one part. The base (white and grey) houses the camera (green) and attaches as the distal phalanx of the finger.}
	\label{fig2}
	\vspace{1em}
	\begin{tabular}{@{}c@{}}
		\includegraphics[width=1\columnwidth,trim={50 65 70 95},clip]{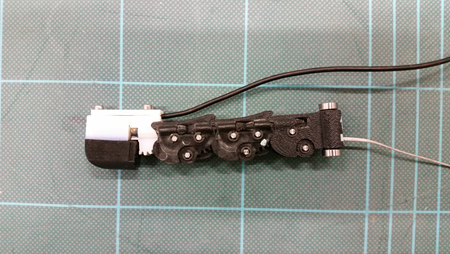}\\
		\includegraphics[width=1\columnwidth,trim={40 85 60 75},clip]{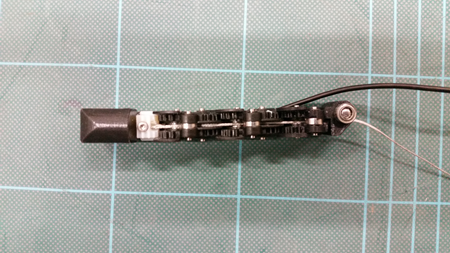}\\
		\includegraphics[width=1\columnwidth,trim={60 95 80 65},clip]{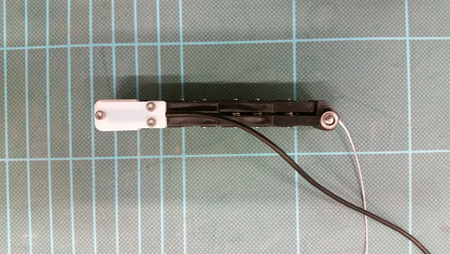}
	\end{tabular}
	\caption{Side, top and bottom view of the integrated tactile fingertip. The single output cable connects to a PC via USB.}
	\label{fig3}
	\vspace{-1em}
\end{figure}

\subsection{Pisa/IIT SoftHand}
\label{sec:3a}

The Pisa/IIT SoftHand is an anthropomorphic robotic hand of similar size to an adult human hand~\cite{catalano_adaptive_2014}. It has 19 joints of which 5 are simple revolute joints that implement the adduction/abduction movement of each finger, and the remaining 14 are compliant rolling-contact element joints. Tendons run from the palm base through all the fingers. The geometry of the hand's bottom part is designed to enable easy connection with standard mechanical interfaces, with a completely self-contained construction such that motors, electronics and sensors are on-board. The hand is open source and available at: \url{www.naturalmachinemotioninitiative.com}.

Each SoftHand digit has a base and three phalanges, based on a repeating arrangement of 3 pulleys that the tendon passes through (technical details in~\cite{catalano_adaptive_2014,santina_toward_2018}). The base is a phalanx modified to end in a revolute joint that attaches to the palm. In the original SoftHand, the distal phalanx is modified into a (non-sensitive) fingertip that contains a pulley to route back the tendon. The interphalangeal soft roll-articular joints each comprise two coupled rolling cams ($6.5\,$mm radius) with gear teeth held together by elastic ligaments.  

To facilitate tactile sensor integration, this work used a non-standard version of the SoftHand with a tendon in each finger that ends on a retainer. In contrast, the original SoftHand has a tendon that runs in a loop from the palm, up and down each finger via a pulley at the tip, then back to the palm. The use of tendon that ends on a retainer made it easier to reroute the terminus of the tendon closer to the distal joint, leaving the internal volume of the distal phalanx free for modification to house the tactile sensor. 

\begin{figure}[t]
	\centering
	\begin{tabular}{@{}c@{}}
		\vspace{.2em}\hspace{-.4em}			
		\includegraphics[width=0.5\columnwidth,trim={0 0 0 0},clip]{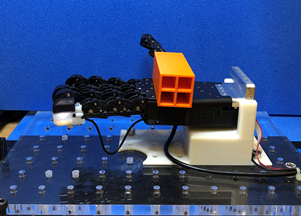}
		\includegraphics[width=0.5\columnwidth,trim={0 0 0 16},clip]{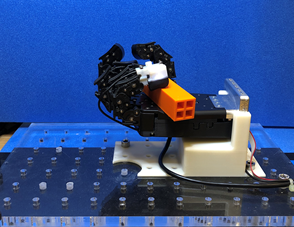} \\	
		\includegraphics[width=0.5\columnwidth,trim={100 0 150 50},clip]{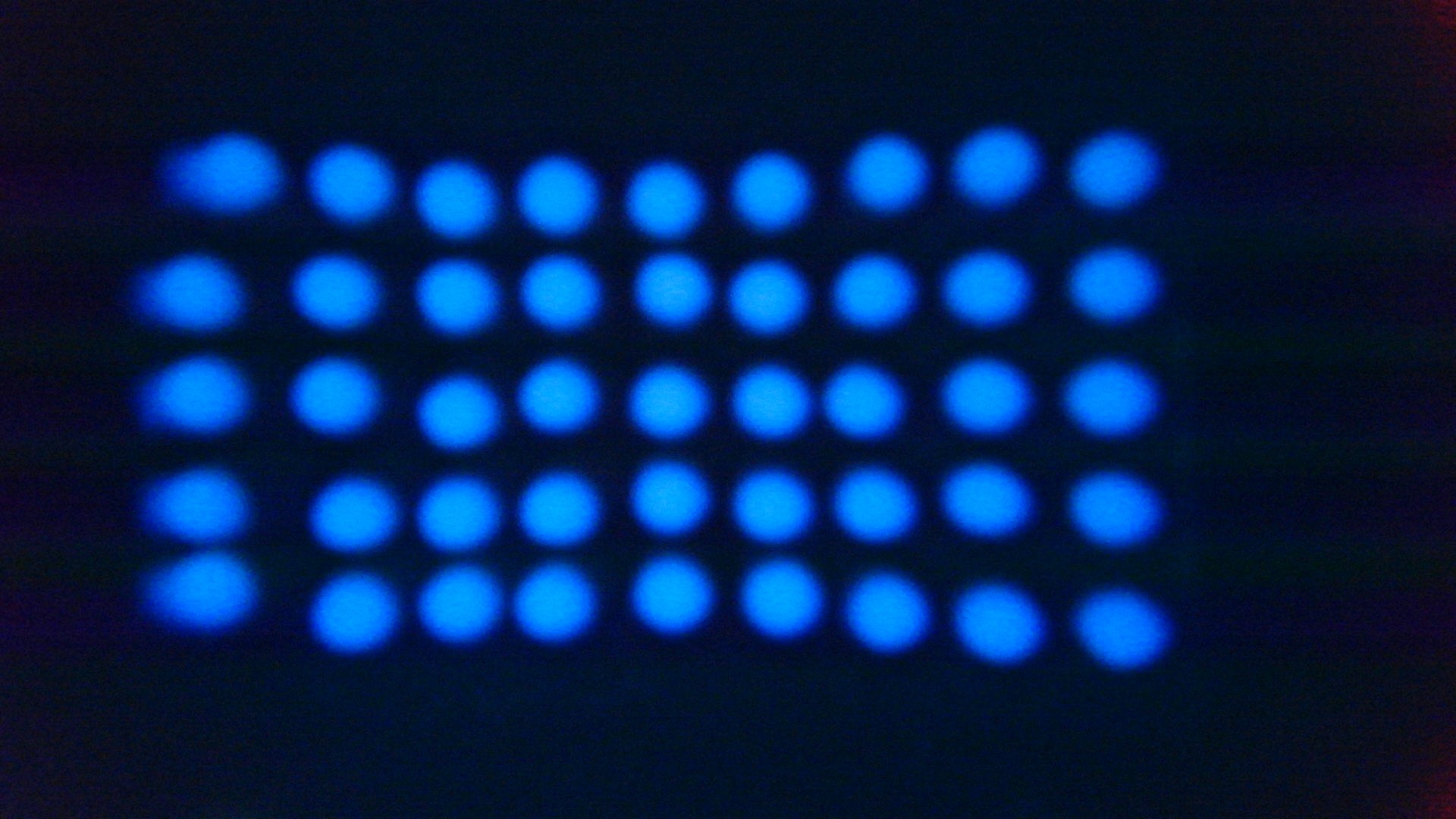}
		\includegraphics[width=0.5\columnwidth,trim={100 0 150 50},clip]{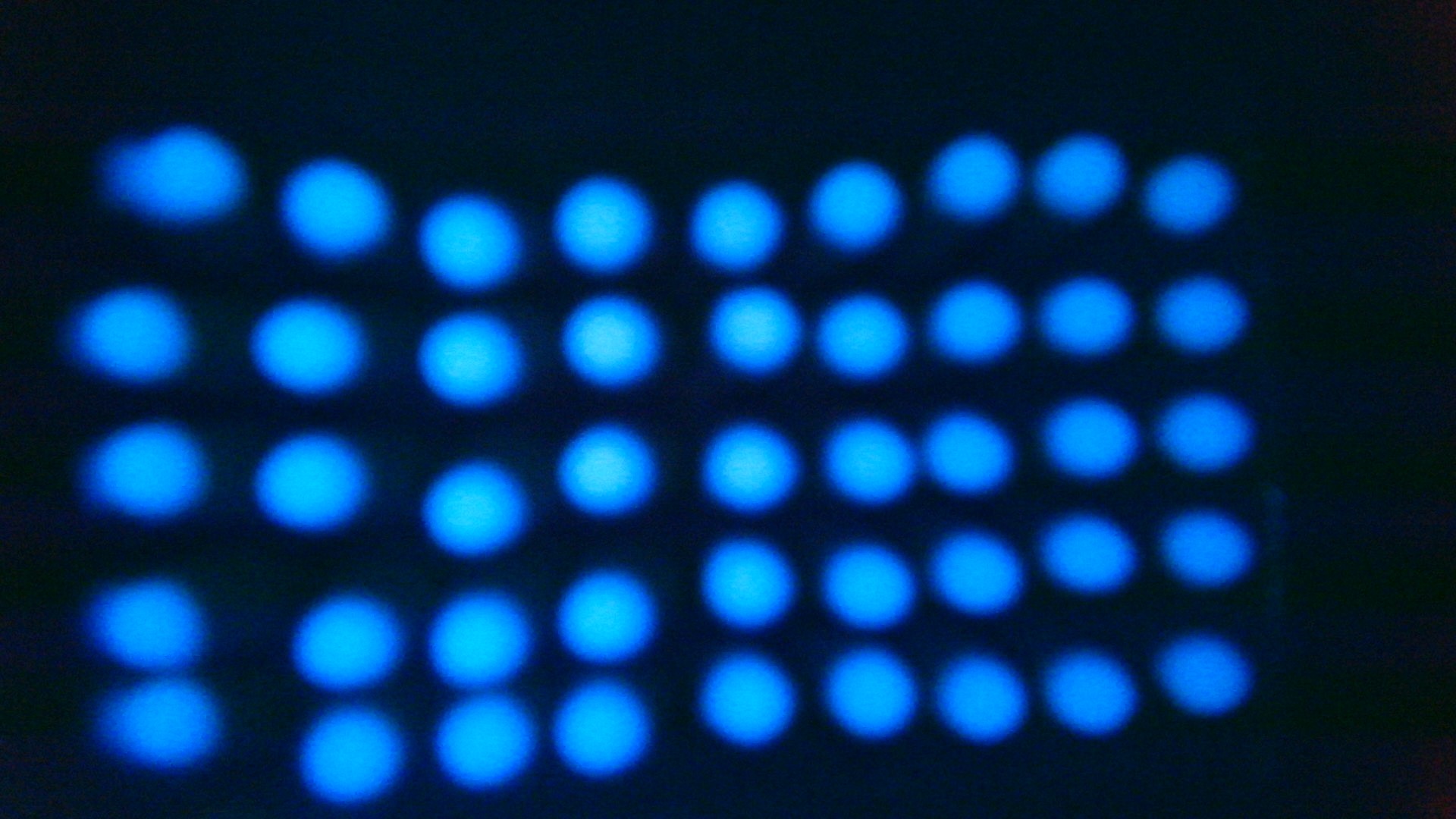} \\
		\includegraphics[width=0.5\columnwidth,trim={0 5 0 10},clip]{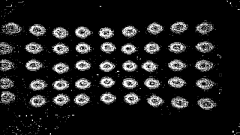}
		\includegraphics[width=0.5\columnwidth,trim={0 5 0 10},clip]{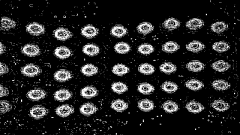} \\
	\end{tabular}
	\caption{Tactile images acquired from the undeformed sensor (left panels) and when the SoftHand is holding an object (right panels). The middle row shows the high-definition sensor output and the bottom row pre-processed (subsampled and thresholded) tactile images.}
	\label{fig4}
	\vspace{-1em}
\end{figure}

\subsection{Tactile sensor-hand integration}
\label{sec:3b}

The tactile fingertip is adapted from an optical biomimetic tactile sensor developed in Bristol Robotics Laboratory: the BRL TacTip. The integration with the SoftHand is facilitated by~\cite{ward-cherrier_tactip_2018}: (i) the use of multi-material 3D-printing to fabricate both the soft and hard components of the fingertip, and (ii)~a modular design in which the tactile fingerpad is separate piece from the base of the sensor that holds the camera. Together these mean that the design and testing can be rapidly iterated to converge on a solution for the integration. 

As each digit of the SoftHand is identical, the tactile sensor can be used on any of the 5 fingers. In this work, we report only on one sensor integrated into the fifth digit, chosen mainly because it is slightly easier to access for testing. 

The integration required several major changes to both the TacTip and the distal phalanx of the SoftHand:
\begin{itemize}
	\item The tendon was rerouted to end on a retaining bolt close to the soft roll-articulator joint, from its original location at the tip of the finger (bolt visible in \autoref{fig3}, middle). 
	\item One side of the soft roll-articular joint was reproduced and extended into a hollow fingertip base for the distal phalanx, which holds the camera and gives a mount for the tip/tactile skin (\autoref{fig2}, middle; \autoref{fig3}, top).  
	\item A backing plate was added to protect the camera electronics, provide strain relief to the cable and result in a neat profile (\autoref{fig2}, right; \autoref{fig3}, bottom panel).
	\item The design was based around housing a miniature camera: the Model SYD (Misumi Electronics) of length $13.5\,$mm, width $6\,$mm and depth $4.9\,$mm, including the lens and electronic component (\autoref{fig2}, to scale).
	\item The camera lens has a $70\,$deg field of view, which constrained the depth of the fingerpad to be at least $10\,$mm for all pins to be viewable.
	\item The camera module has its own LED lighting that we use to illuminate inside the fingerpad, powered by a single USB3 cable connection to a PC. 
	\item The camera has maximal resolution of 1080p  at $60\,$fps, although we acquire images at lower resolution to reduce image processing and storage. 
\end{itemize}

Other design aspects were evolved from versions of the TacTip integrated into robot hands~\cite{ward-cherrier_model-free_2017,pestell_sense_2019,james_tactile_2020,ward-cherrier_miniaturised_2020}:
\begin{itemize}
	\item The fingerpad is fabricated as one part with a multi-material 3D-printer (Stratasys Objet), comprising a rigid rim (Vero White) and a compliant skin (Tango Black+), with the rim slotting onto the base section.
	\item The inside of the skin extends into a rectangular array of $5\times9$ pins ($1\,$mm length and $0.75\,$mm diameter). Markers on the end of these pins are printed in Vero White. 
	\item A clear acrylic sheet ($1\,$mm thick) is glued into the rim to give a small cavity, which is filled with an optically-clear silicone gel (RTV27905, Techsil UK) that gives the fingerpad a compliance similar to a human fingertip.
\end{itemize}

\subsection{Tactile data acquisition and processing}
\label{sec:3c}

Deformation of the tactile sensing pad is imaged with the internal camera at its native resolution of 1080p, then adaptively thresholded with a Gaussian filter (width 39, mean 0 pixels) and subsampled/cropped to $(240\times135)$-pixel grey-scale images (\autoref{fig4}). All image acquisition and processing was carried out in Python OpenCV. The tactile image data was used in two ways:

\subsubsection{Contact deformation measurement} A simple yet robust measure of the difference in tactile images can be found from the Structural Similarity Index Measure (SSIM)~\cite{luo_vitac_2018}, which can be used to measure contact deformation by comparing a tactile image against a non-deformed reference image~\cite{james_tactile_2020}. Here we use 
\begin{equation}
e_{\rm SSIM}(I)=1-{\rm SSIM}(I,I_{\rm ref})
\end{equation} 
as a measure of the deformation of image $I$ compared with the reference image $I_{\rm ref}$, with ${\rm SSIM}$ implemented using Python SciKit-Image and computed from the local means, variances and cross-covariance of the two images~\cite{wang_image_2004}. The RMS pixel intensity change was also explored, but it was not useful as it saturated at small deformations. The SSIM-based deformation measure changed gradually as the contact intensified, making it suitable for use in a feedback controller.

\begin{figure}[t]
	\centering
	\begin{tabular}{@{}cc@{}}
		\begin{minipage}[b]{0.45\columnwidth}
			\begin{tabular}{@{}c@{}}
				\includegraphics[width=\columnwidth,trim={300 190 310 60},clip]{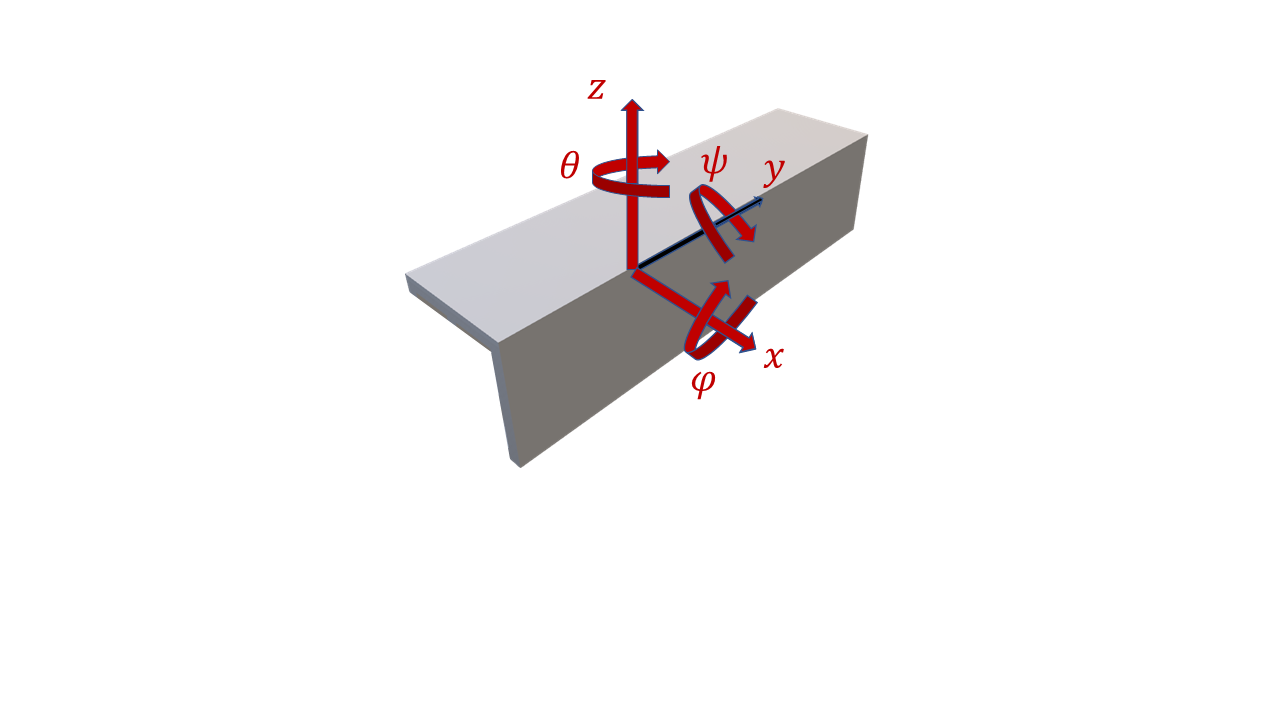} 
			\end{tabular}
		\end{minipage} &
		\small
		\begin{minipage}[t]{0.6\columnwidth}
			\begin{tabular}{@{}c|c@{}c@{}}
				\textbf{Pose} & \textbf{Range} & \textbf{Perturbation}\\ 
				\hline
				$x$ & [-6,6]\,mm & [-2,2]\,mm \\ 
				$y$ & - & [-2,2]\,mm \\ 
				$z$ & [0,3]\,mm& [-1,1]\,mm  \\ 
				$\phi$ & [-5,5]\,deg& [-2,2]\,deg  \\
				$\psi$ & [-10,10]\,deg& [-2,2]\,deg  \\
				$\theta$ & [-45,45]\,deg& [-2,2]\,deg  \\  
			\end{tabular}
		\end{minipage}
	\end{tabular}
	\caption{Pose ranges used as labels to train/test the neural network along with unlabelled shear perturbations of the contact for aiding generalization~\cite{lepora_optimal_2020}.}
	\vspace{-1em}
	\label{fig5}
\end{figure}

\begin{table}[b]
	\vspace{-1em}
	\begin{tabular}{lc}
		\textbf{Hyperparameters} &  \textbf{Optimized values} \\ 
		\hline
		\# convolutional hidden layers, $N_{\rm conv}$ & 5 \\ 
		\# convolutional kernels, $N_{\rm filters} $ & 256 \\ 
		\# dense hidden layers, $N_{\rm dense}$ & 1 \\
		\# dense hidden layer units, $N_{\rm unit}$ & 256 \\ 
		hidden layer activation function & ReLU \\ 
		dropout coefficient & 0.02 \\ 
		L1-regularization coefficient & 0.0001 \\ 
		L2-regularization coefficient & 0.0005 \\ 
		batch size & 16 \\ 
	\end{tabular}
	\caption{Neural network and learning hyperparameters.}
	\label{tab1}
	\vspace{-1.5em}
\end{table}

\subsubsection{Edge pose estimation} A complementary use of the tactile images is to estimate the 3D pose from the edge of a contacted object relative to the sensor (\autoref{fig5}), and thus estimate the pose of an object held in hand. We have recently published a comprehensive paper on how to apply convolutional neural networks to pose estimation~\cite{lepora_optimal_2020} which we refer to for the details of the deep learning used here. 

The principal difference from our previous work is that the tactile fingertip was trained while mounted on the SoftHand. This was arranged with a custom test platform comprising a mount that immobilises the hand and fingertip, combined with a test stimulus mounted on a robot arm (photo in \autoref{fig8}). The stimulus was then contacted repeatedly against the fingertip to gather the large amount (10,000 contacts) of labelled data needed to train and test the deep neural network. 

We used a result from ref.~\cite{lepora_optimal_2020} that the pose estimation can be improved by introducing motion-dependent shear into the training data collection. Thus, each sample of data had a random labelled pose and a random unlabelled shear perturbation (ranges in \autoref{fig5}). The optimized network hyperparameters are reported in \autoref{tab1}, with the training implemented in the Python Keras library using a Titan Xp GPU (12Gb memory). Once trained, the deep neural network was deployed on the CPU of a standard laptop. 

\begin{figure}[t]
	\centering
	\begin{tabular}{@{}c@{}}
		\includegraphics[width=1\columnwidth,trim={175 150 165 120},clip]{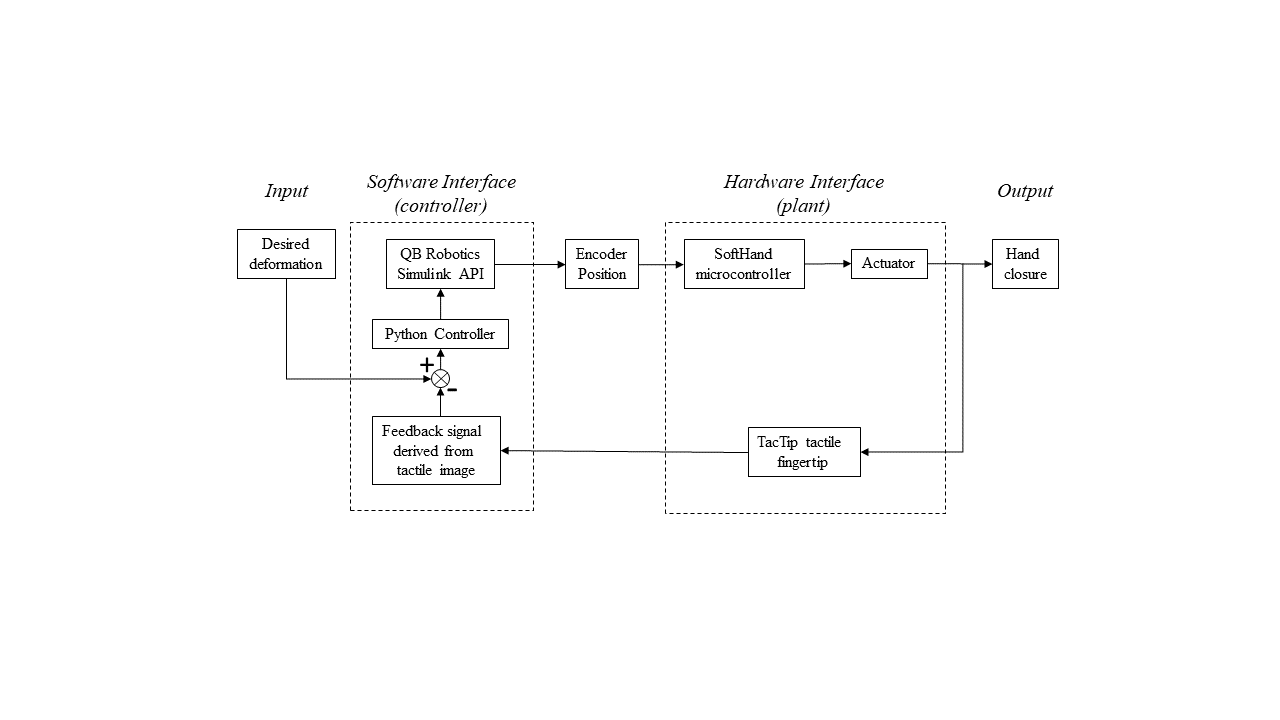}
	\end{tabular}
	\caption{SoftHand controller with feedback derived from the tactile image.}
	\vspace{-1em}
	\label{fig6}
\end{figure}

\subsection{Tactile SoftHand control}
\label{sec:3d}

The SoftHand control software (\autoref{fig6}) combines a feedback loop and tactile processing pipeline implemented in Python with a Simulink interface (qbmove, distrubuted by qb Robotics) that sends motor commands to the robot hand. The Simulink interface is built upon a C++ API for controlling the motor in the hand, compiled into MEX files for consolidation into Simulink blocks. Real-time communicateion between Python and Simulink used the MATLAB Engine library. 

In the Python control loop, the actuator is driven by incrementally changing the set-point for the encoder on the motor of the SoftHand, rather than directly driving the actuator (because this functionality was not part of the qbmove Simulink interface). Hence, the controller does not directly use the motor encoder position as feedback, but instead increments the set-point and then waits for that point to be reached before the next Python control loop can begin. This process functioned effectively for the experiments presented here, but does lead to a slow cycle time between the processed tactile images and the hand control (typically 100-200\,ms). 

A proportional-gain feedback controller was implemented based on the SSIM contact deformation defined above, with a change in motor command $u(t)$ sent to the hand
\begin{equation}
\Delta u(t) = g_{\rm P} \left(e_{\rm SSIM}(I(t)) - r\right),
\end{equation}
where $I(t)$ is the tactile image at time $t$, $g_{\rm P}=100$ is a (hand-tuned) proportional gain and $r=0.7$ the set-point. A second controller was considered using the $z$-component of the edge pose estimation from the neural network, 
\begin{equation}
\Delta u(t) = g_{\rm P} \left(z(t) - r_z\right),
\end{equation}
with the set point $r_z$ in the range 0-3\,mm. The motor command has range $0\leq u\leq u_{\rm max}$ with $u_{\rm max}=19000$.

\section{RESULTS}
\label{sec:4}

\begin{figure*}[t]
	\centering
	\begin{tabular}{@{}c@{}}
		\vspace{.2em}\hspace{-.4em}			
		\includegraphics[width=0.5\columnwidth,trim={100 10 0 30},angle=0,clip]{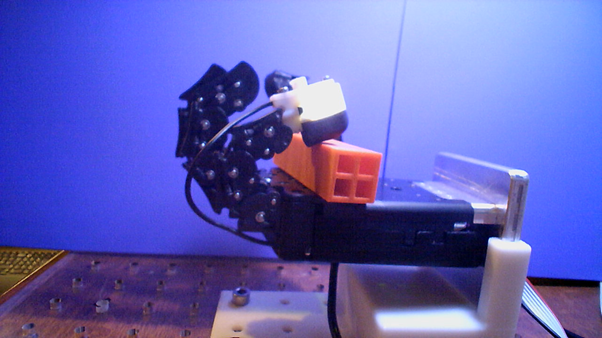}
		\includegraphics[width=0.5\columnwidth,trim={100 10 0 30},angle=0,clip]{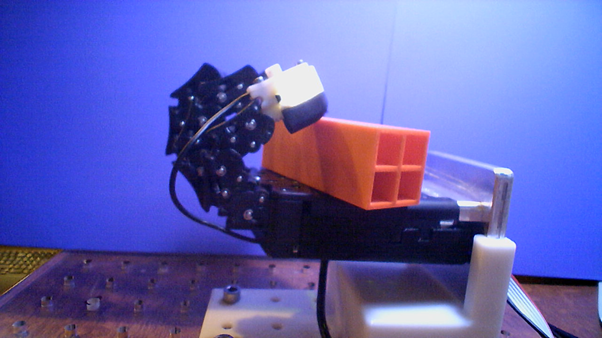} 
		\includegraphics[width=0.5\columnwidth,trim={100 10 0 30},angle=0,clip]{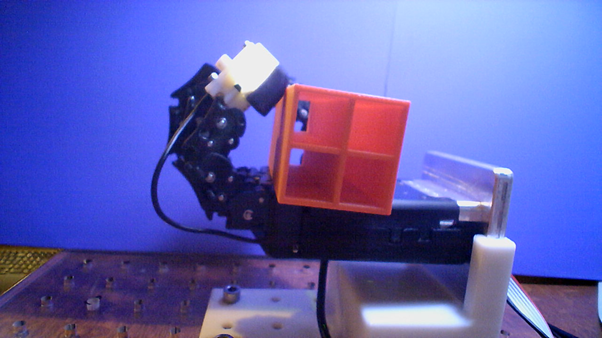}
		\includegraphics[width=0.5\columnwidth,trim={100 10 0 30},angle=0,clip]{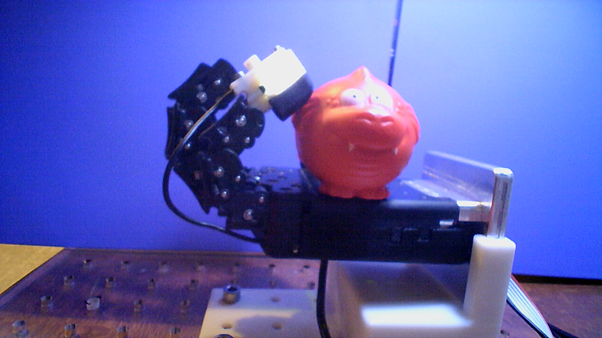} 
		\\	
		\includegraphics[width=0.5\columnwidth,trim={0 0 0 10},angle=0,clip]{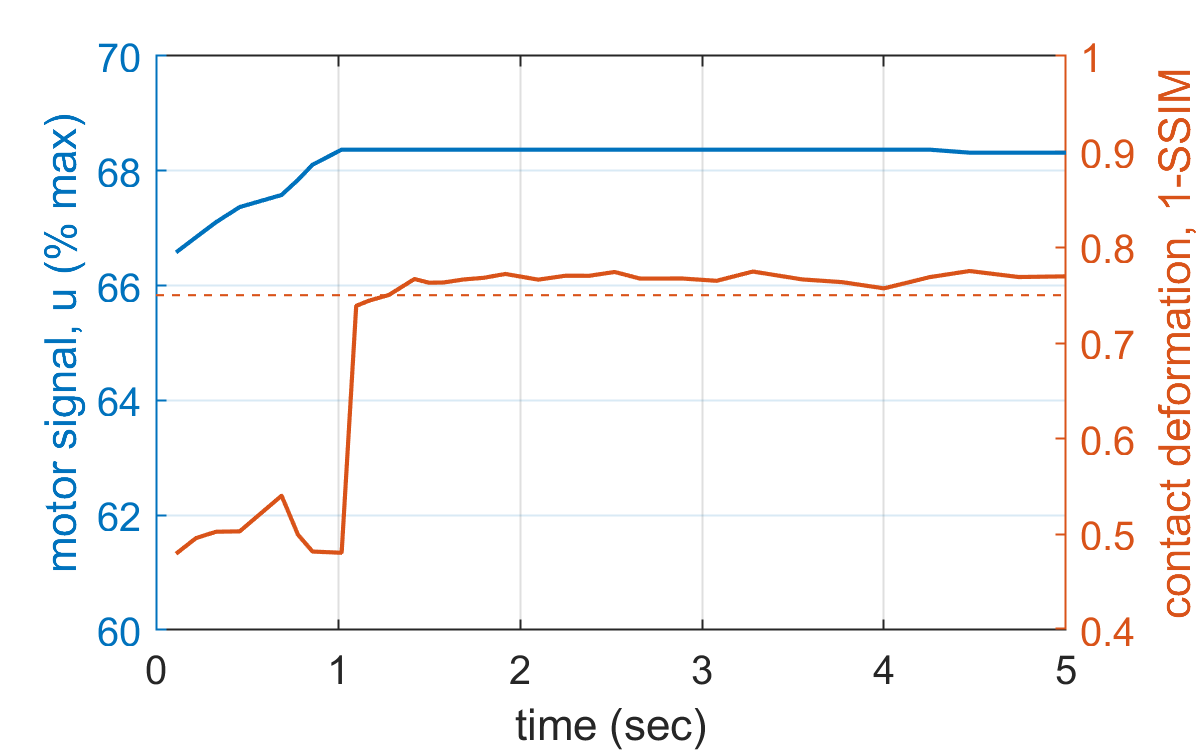}
		\includegraphics[width=0.5\columnwidth,trim={0 0 0 10},angle=0,clip]{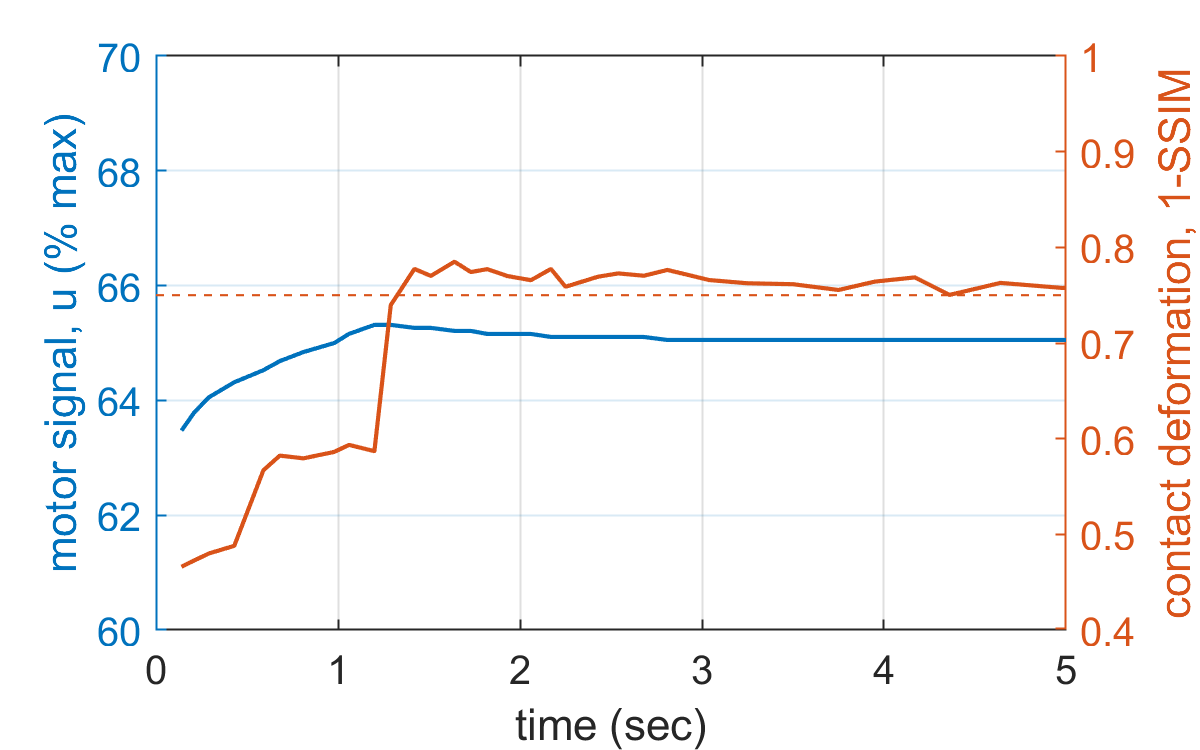} 
		\includegraphics[width=0.5\columnwidth,trim={0 0 0 10},angle=0,clip]{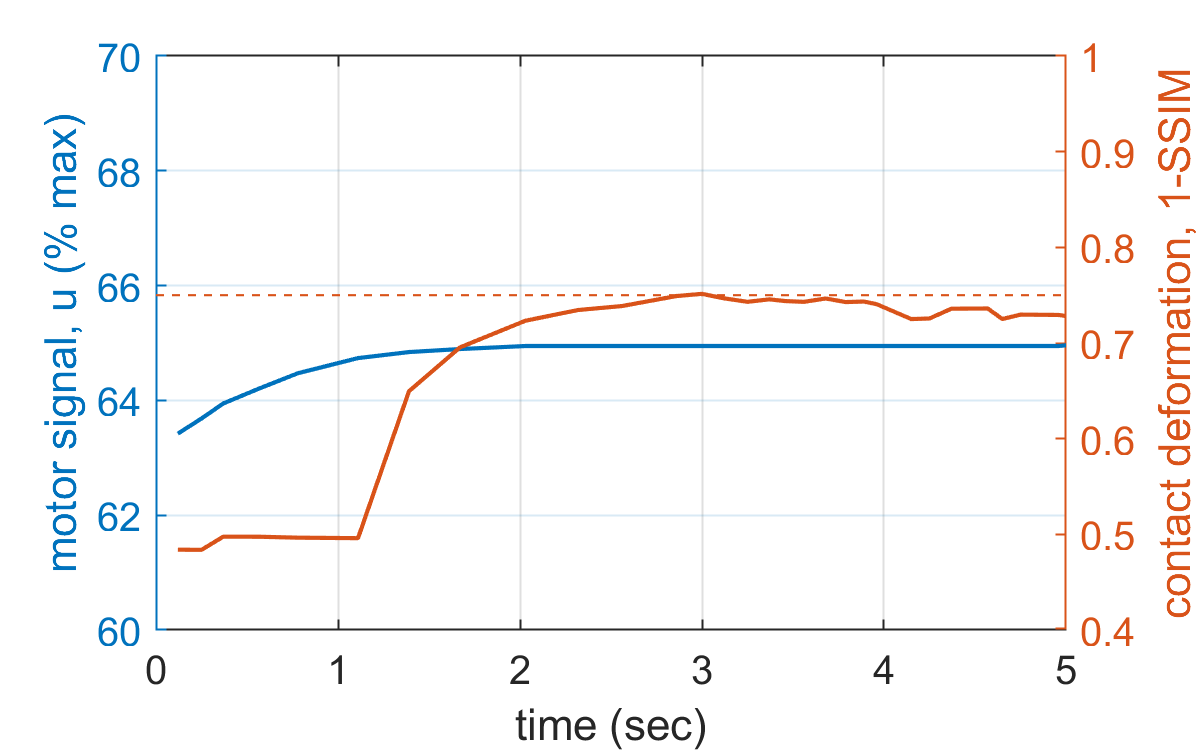}
		\includegraphics[width=0.5\columnwidth,trim={0 0 0 10},angle=0,clip]{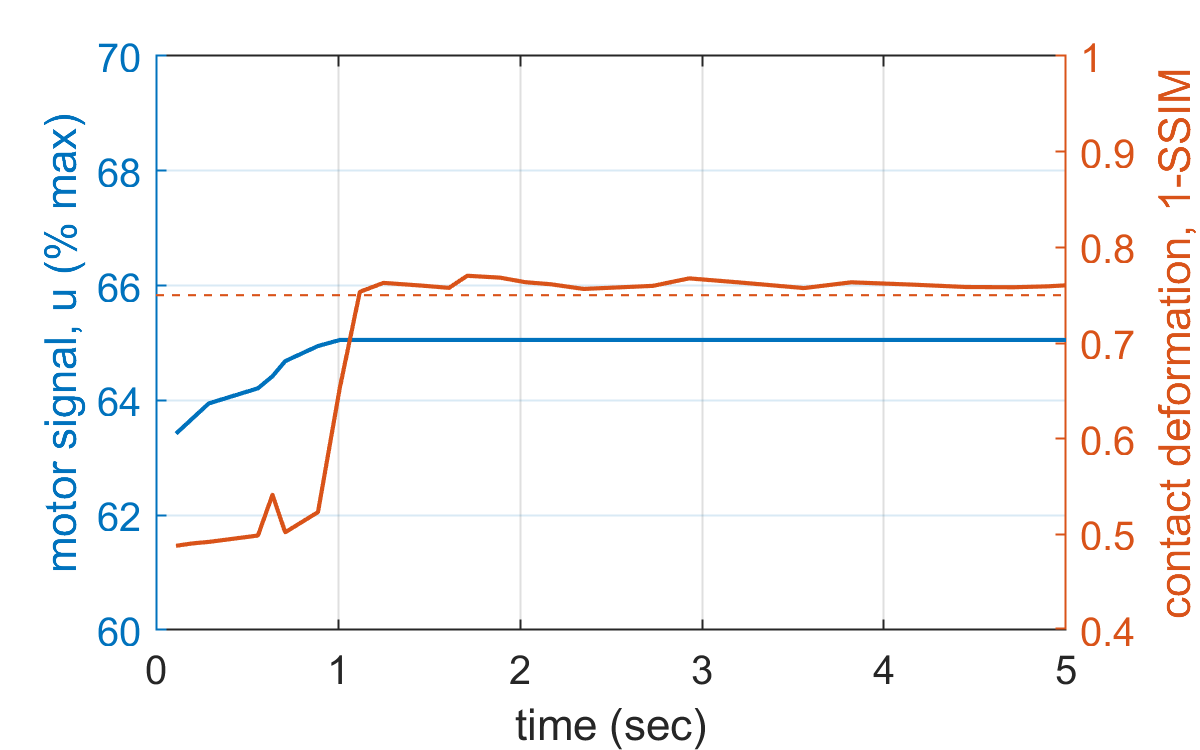} 
		\\
	\end{tabular}
	\caption{\footnotesize Experiment 1 tests the control of light touch on four objects: 3 square prisms (20\,mm, 30\,mm, 40\,mm) and an irregular soft object. The controller maintain the SSIM contact deformation at a set point $r=0.7$ deformation from no contact. A video of this experiment is provided with this paper.}
	\label{fig7}
	\vspace{-1em}
\end{figure*}

\subsection{Experiment 1: Adaptive grasp closure from touch}
\label{sec:4a}

\begin{figure}[b!]
	\vspace{-1em}
	\centering
	\includegraphics[width=0.8\columnwidth,trim={0 50 0 40},clip]{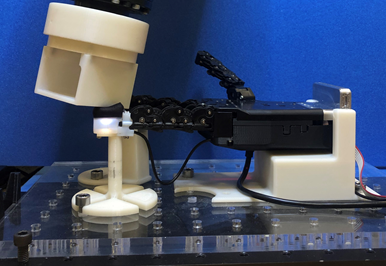}\\
	\includegraphics[width=1\columnwidth,trim={45 0 40 0},clip]{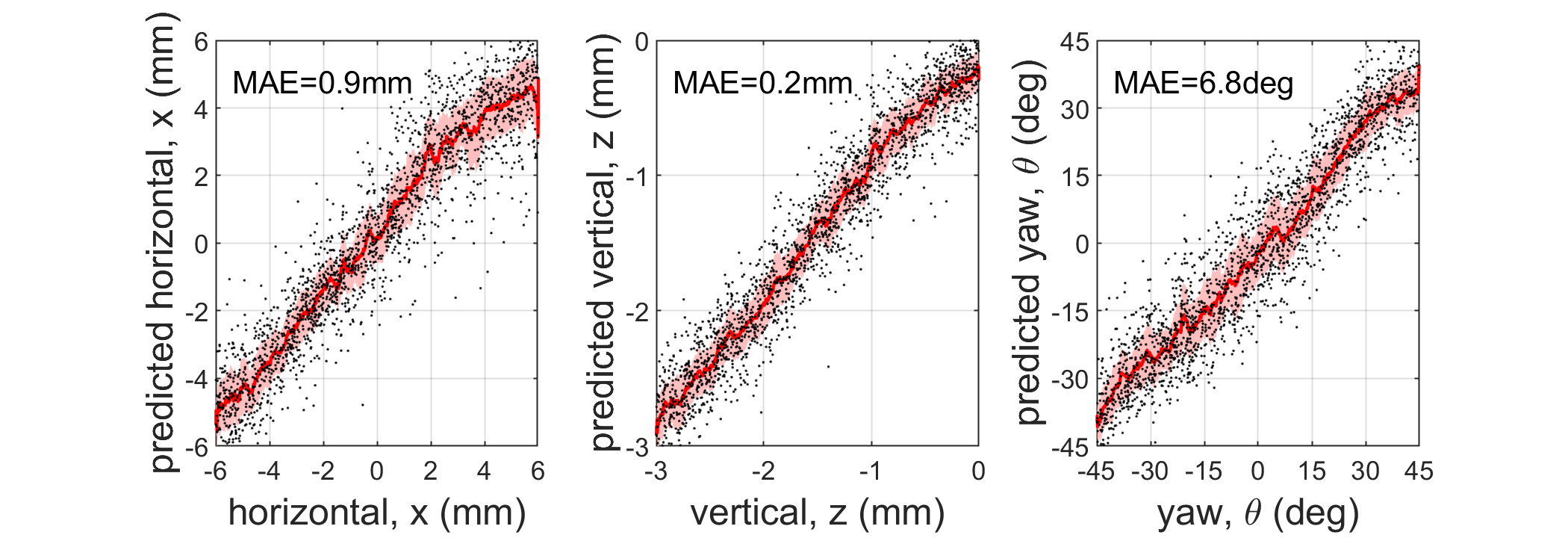}\\
	\includegraphics[width=0.67\columnwidth,trim={20 0 25 0},clip]{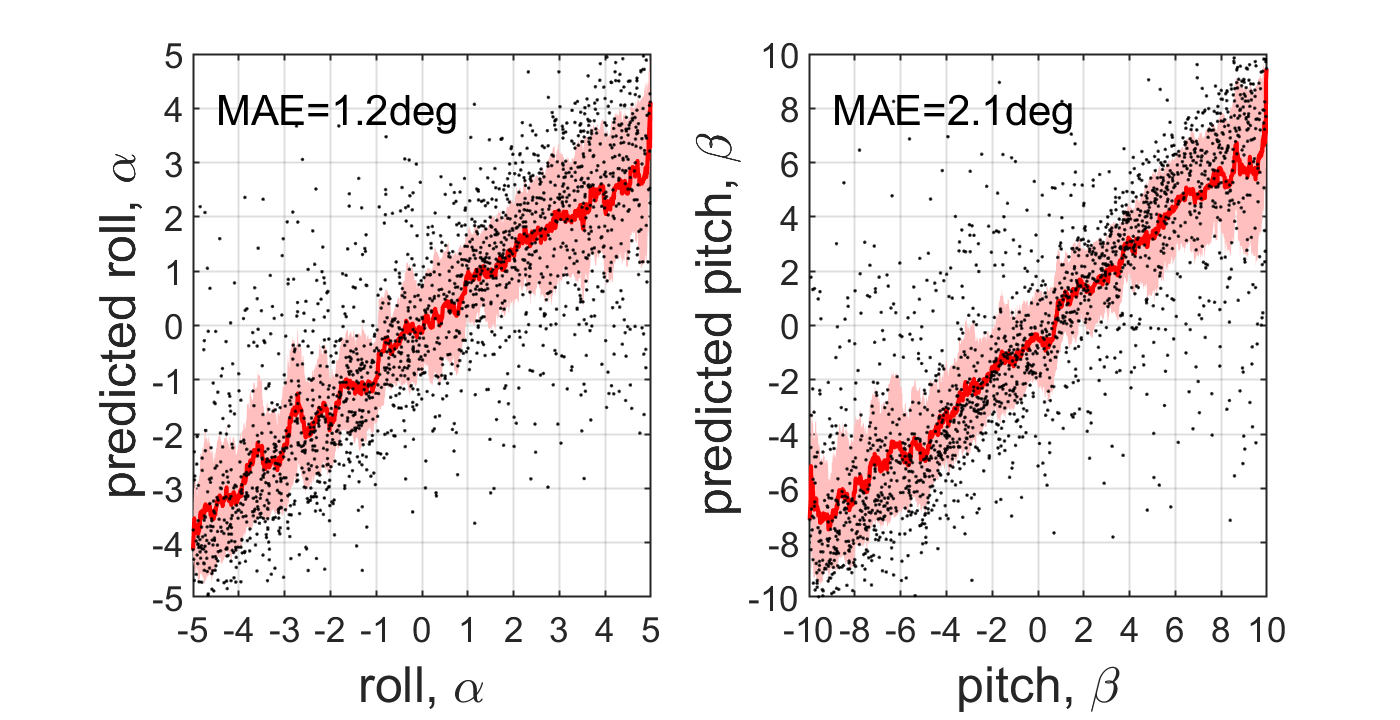}
	\caption{Experiment 2 tests the estimation of object edge pose. The PoseNet performance is shown along with smoothed predictions (red; 50-sample moving average) and mean absolute errors (MAE) assessed as in Ref.~\cite{lepora_optimal_2020}.}
	\label{fig8}
\end{figure}

The first experiment tests the performance of the tactile SoftHand using the feedback controller in \autoref{fig6} to maintain a light touch as the hand closes around an object. As described in the Methods, we use a proportional-gain feedback controller on the set-point of the actuator, with feedback signal a measure of contact deformation estimated from comparing the tactile image against an undeformed reference using the structural similarity SSIM. 

The adaptation of the grasp closure was tested with 4 distinct objects and found to work well with no failure cases under various object poses (\autoref{fig7}). The grasp closure was controlled to be sufficiently light that there was little if any movement of the objects as the finger contacted them. This is evident in~\autoref{fig7} by the motor coming to a halt as the SSIM control signal reaches its set point. 

In our view, the main use of this capability is to make an initial light contact that can then be adapted further using other information from the tactile feedback about the object and its pose. Therefore, we now consider the quality of tactile information available from the sensor. 

\subsection{Experiment 2: Tactile perception of object edge pose}
\label{sec:4b}

\begin{table}[b!]
	\vspace{-1em}
	\begin{tabular}{cc@{}|@{}cccc@{}|@{}ccc@{}}
		\textbf{Pose} &&& \multicolumn{2}{c}{\textbf{SoftHand TacTip}} &&& \multicolumn{2}{c}{\textbf{Original TacTip~\cite{lepora_optimal_2020}}} \\ 
		\textbf{component} &&& \textbf{MAE} & \textbf{range} &&& \textbf{MAE} & \textbf{range}\\ 
		\hline
		horizontal, $x$ &&& 0.9\,mm & 12\,mm &&& 0.6\,mm & 10\,mm \\ 
		vertical, $z$ &&& 0.2\,mm& 3\,mm  &&& 0.2\,mm& 4\,mm  \\ 
		roll, $\phi$ &&& 1.2\,deg& 10\,deg  &&& 1.4\,deg& 30\,deg  \\
		pitch, $\psi$ &&& 2.1\,deg& 20\,deg  &&& 1.6\,deg& 30\,deg  \\
		yaw, $\theta$ &&& 6.9\,deg& 90\,deg  &&& 6.4\,deg& 90\,deg  \\  
	\end{tabular}
	\caption{PoseNet prediction performance}
	\label{tab2}
\end{table}

To test the quality of the tactile sensing from the integrated fingertip, we assess its performance on an edge pose estimation task that has been examined in detail with the standard design of TacTip sensor~\cite{lepora_optimal_2020}. This experiment is both useful in itself, to develop a trained model for estimating the pose of held objects, and also serves as benchmark on the performance of the integrated optical tactile sensor. 

Data were gathered for this test by using a robot arm to bring a test object into contact with the tactile fingertip while the finger was held immobile (\autoref{fig8}, top; also \mbox{Sec.~\ref{sec:3c}}). A range of object poses were considered that varied in all dimensions except along the edge (Table in \autoref{fig5}), with shear perturbations applied to the contact. A `PoseNet' convolutional neural network was trained to regress pose over the labelled tactile images (\autoref{sec:3c}). The training involved Bayesian optimization over the network architecture and hyperparameters~\cite{lepora_optimal_2020} (optima reported in ~\autoref{tab1}). 

Pose estimation was then assessed on a distinct test set. Plots of the predictions versus labels show good performance for three $(x,z,\theta)$ pose components and fair performance for the other angular $(\phi,\psi)$ components relative to their ranges (Figure~\ref{fig8}). We emphasise that these predictions are after a random unknown shear, which makes the regression more difficult because of the latent variable but aids generalisation.

In comparison with the standard design of TacTip sensor used in Ref.~\cite{lepora_optimal_2020}, the accuracies were similar for the $(x,z,\theta)$ pose components (Table~\ref{tab2}). Ranges of the angular components $(\phi,\psi)$ were smaller that those used with the standard TacTip to avoid damage as the stimulus tilts into the hand. 

\begin{figure}[t]
	\centering
	\begin{tabular}{@{}c@{}}
		\includegraphics[width=\columnwidth,trim={30 0 10 0},clip]{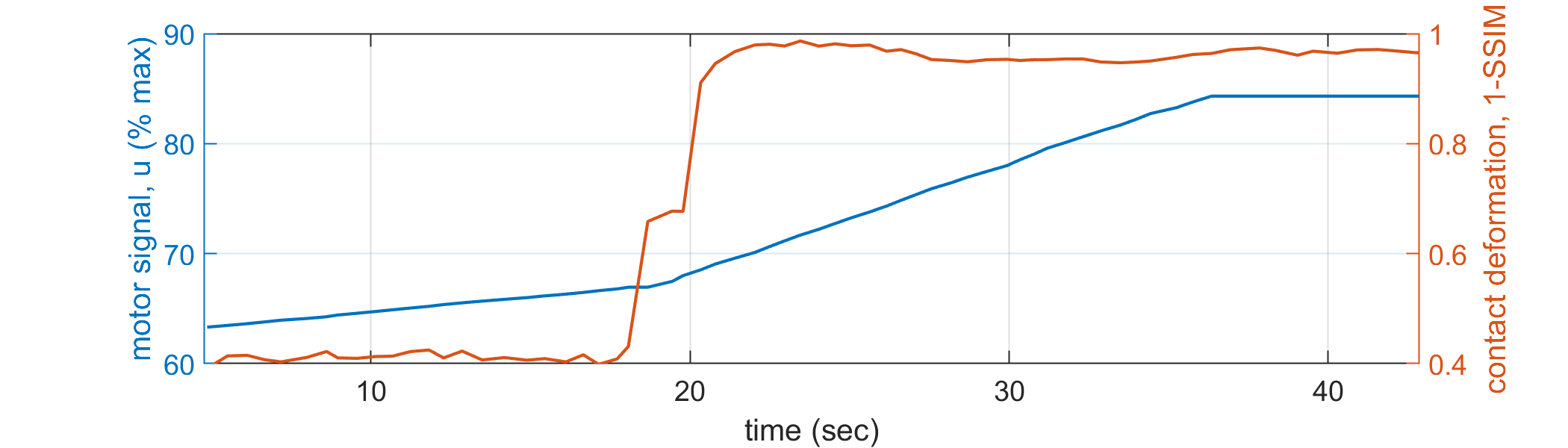}\\
		\includegraphics[width=\columnwidth,trim={40 0 20 0},clip]{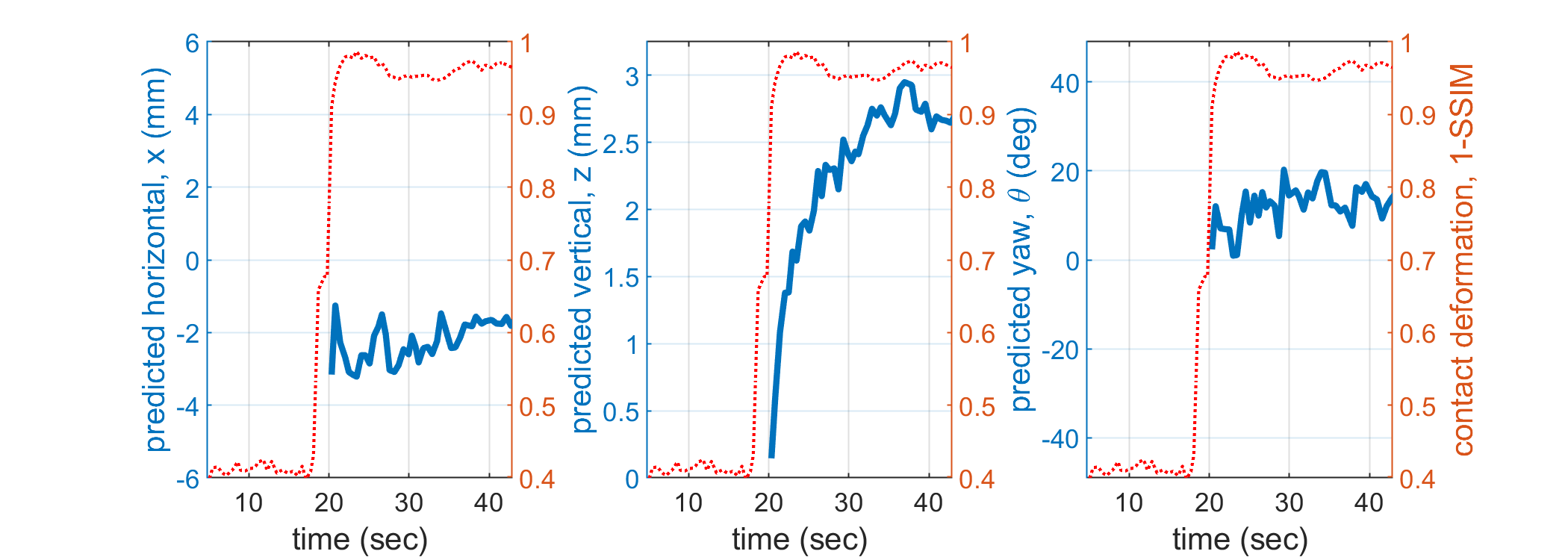}\\
	\end{tabular}
	\vspace{-0.75em}
	\caption{Experiment 3A closes the hand using SSIM feedback control then ramps the motor signal. Edge pose is estimated from the tactile images.}
	\label{fig9}
	\vspace{0.5em}
	\centering
	\begin{tabular}{@{}c@{}}
		\includegraphics[width=\columnwidth,trim={30 0 10 0},clip]{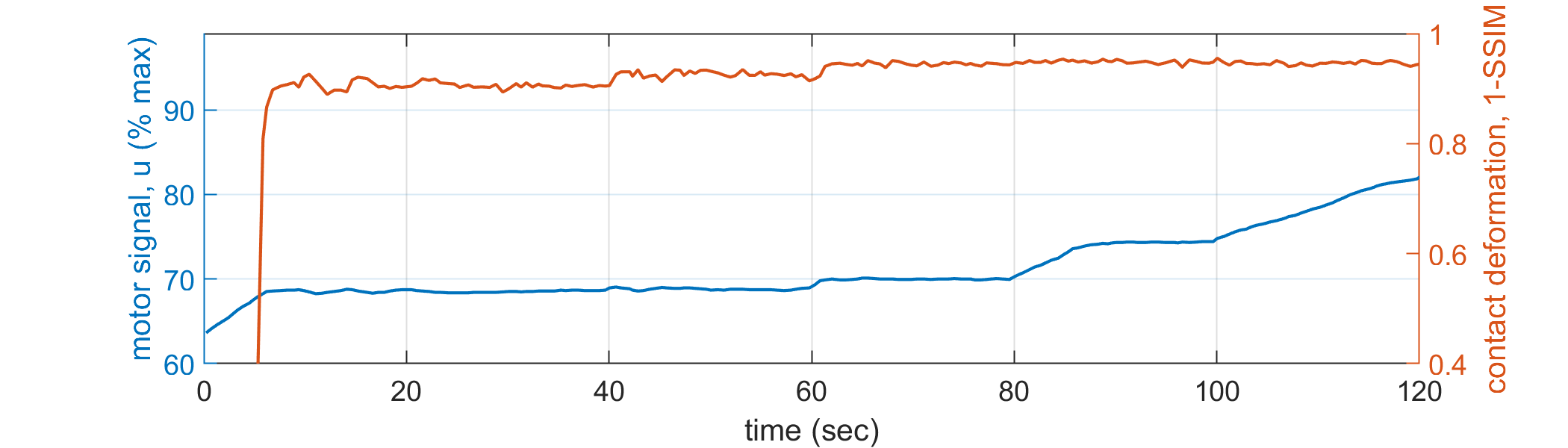}\\
		\includegraphics[width=\columnwidth,trim={40 0 20 0},clip]{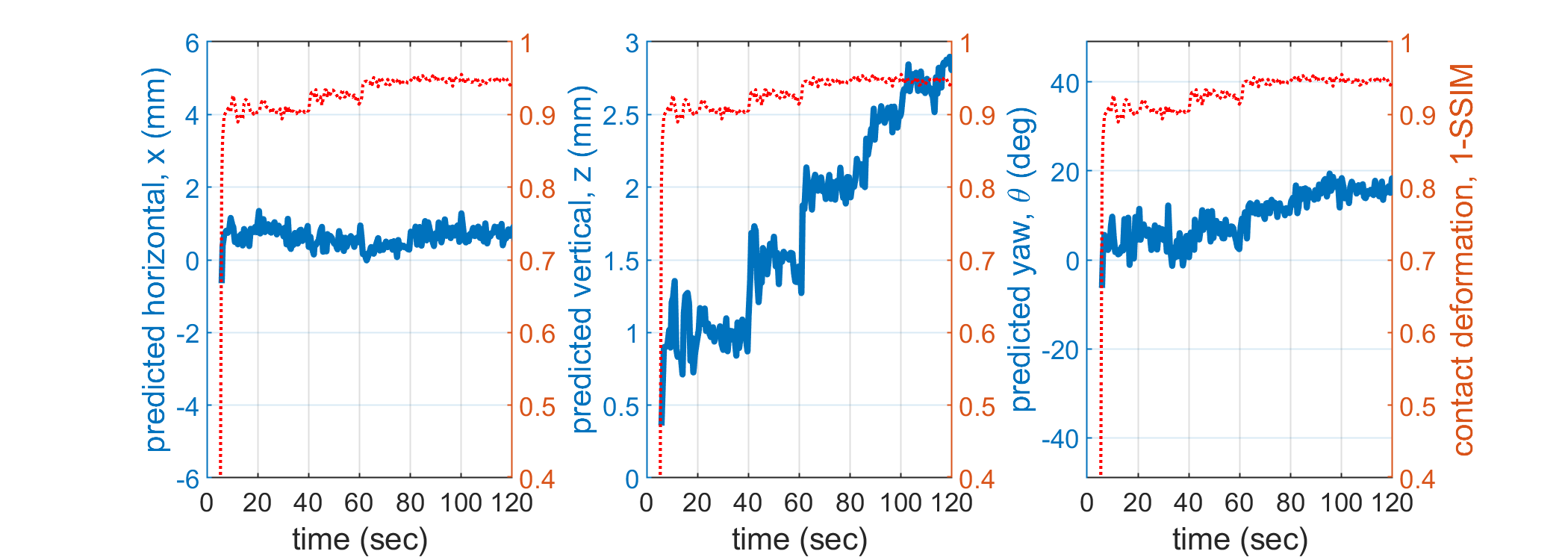}\\
	\end{tabular}
	\vspace{-0.75em}
	\caption{Experiment 3B closes the hand using SSIM feedback control then controls hand closure based on the $z$-pose estimated from the tactile images, with set points 1\,mm, 1.5\,mm, 2\,mm, 2.5\,mm and 3\,mm every 20 seconds. A video of both experiments 3A,B is provided with this paper.}
	\vspace{-1em}
	\label{fig10}
\end{figure}

\subsection{Experiment 3: Object pose during adaptive grasp closure}
\label{sec:4c}

Lastly, we combine the adaptive grasp closure and tactile perception capabilities in two experiments: first, ramping the motor signal (by 1\% per second) while estimating object pose, and second, using the estimated pose to control hand closure. In both experiments, we start by closing the hand for 20\,sec using the SSIM contact deformation for a light grasp, using the smallest square prism (20\,mm) as stimulus. 

Note that Experiments 3A,B (\autoref{fig9},\autoref{fig10}) are about $4\times$ slower than Experiment 1 (\autoref{fig7}) because of the deep neural network. Although pose estimates took only $\sim50$\,ms, the computational load caused slower image acquisition for the videos provided to accompany the figures; otherwise, all experiments would run at a similar rate. 

Experiment 3A (\autoref{fig9}) shows that pose estimation remains stable during the test with noise as expected from the offline results in Experiment 2 (\autoref{fig8}). For light contacts, both the SSIM contact deformation and $z$-pose increase together. At medium to strong contacts, the $z$-pose becomes more sensitive than the SSIM, which has a saturating non-linearity. Conversely, for almost no contact, the neural network becomes unreliable (defaulting to a mean prediction). Thus, poses are only shown for $1-{\rm SSIM}>0.45$. 

Experiment 3B (\autoref{fig10}) shows that the $z$-pose is suitable for feedback control to modulate the adaptive grasp closure, by stepping through 5 set points from 1\,mm to 3\,mm. At each set point, the motor signal equilibrates at the value needed to maintain that contact on the fingertip while the other pose estimates remain stable. 



\section{Discussion}

In this paper, we have presented an integration of the BRL tactile fingertip (TacTip) and the anthropomorphic Pisa/IIT SoftHand that is able to finely control its hand closure using tactile feedback. Two measures from the tactile images gave suitable feedback signals for controlling hand closure: (i) a structural similarity index measure (SSIM)~\cite{wang_image_2004} of contact deformation for very light or no contact; and (ii) object edge pose estimation from a convolutional neural network~\cite{lepora_optimal_2020} for light/medium to strong contacts. Hence, the SSIM contact deformation can guide hand closure to an initial contact, followed by pose estimation to adapt that contact or grasp.

This initial study considered tactile sensing from a single fingertip to show the potential for a tactile soft hand. In principle, it should be straightforward to integrate the other four fingertips, as all fingers of the SoftHand have the same design. Another capability to explore is slip correction, as the TacTip is effective for detecting and correcting slip when integrated into a 3-fingered gripper~\cite{james_slip_2020}. The SoftHand used here only has one degree of actuation, but its softness enables greater dexterity by interacting with the environment. Tactile sensing from multiple fingertips could control those interactions or help control soft robotic hands with more degrees of actuation such as the Pisa/IIT SoftHand 2~\cite{santina_toward_2018}. 

In terms of the design, the integrated tactile sensor has a similar size and shape to a human fingertip, with a soft surface compatible with the soft hand design. It is also robust: the same fingertip was used without wear or damage over tens of thousands of contacts and under driving the hand to near its maximal grasping force. This robustness complements that of the SoftHand from their soft designs.  

Future work will encompass a more complete evaluation of the performance in different autonomous exploration and grasping tasks with a real-time implementation of the sensorimotor control and a fully 5-fingered sensorized hand.  



\section*{Acknowledgment} 
We thank Vinicio Tincani and Cristiano Petrocelli (IIT) for the preparation of the hand and CAD, and Elim Kwan for internship work on this project. This work was supported by Leverhulme Trust Research Leadership award on `A biomimetic forebrain for robot touch' (RL-2016-39), the EU's H2020 Programme under grant agreements SOPHIA (871237), ERC Synergy Grant Natural BionicS (810346) and ReconCycle (871352), and and by the Italian Ministry of Education and Research (MIUR) in the framework of the CrossLab project (Departments of Excellence).

\newpage\pagebreak



\bibliographystyle{unsrt}
\bibliography{ICRA2021-Softhand}

\end{document}